\pdfoutput=1

\documentclass[11pt]{article}

\usepackage{acl}

\usepackage{times}
\usepackage{latexsym}

\usepackage[T1]{fontenc}

\usepackage[utf8]{inputenc}

\usepackage{microtype}

\usepackage{inconsolata}
\usepackage{hyperref}
\usepackage[utf8]{inputenc}
\usepackage{url}            
\usepackage{booktabs}       
\usepackage{amsfonts}       
\usepackage{nicefrac}       
\usepackage{microtype}      
\usepackage{xcolor}         
\usepackage{graphicx}
\usepackage{subfigure}
\usepackage{amsmath}
\usepackage{multirow}
\usepackage{algorithm, algorithmic}
\usepackage{caption}
\usepackage{pifont}
\usepackage{wrapfig,lipsum,booktabs}
\usepackage{lineno}
\usepackage{colortbl}
\usepackage{color}
\usepackage{longtable}
\definecolor{Blue2}{RGB}{235, 245, 250}

\title{Small LLMs Are Weak Tool Learners: A Multi-LLM Agent}

\author{Weizhou Shen\textsuperscript{\rm 1}, Chenliang Li\textsuperscript{\rm 2}, Hongzhan Chen\textsuperscript{\rm 1}, Ming Yan\textsuperscript{\rm 2}\thanks{$\;\;$Corresponding authors.},\\
        \textbf{Xiaojun Quan}\textsuperscript{\rm 1}\textsuperscript{$*$}, \textbf{Hehong Chen}\textsuperscript{\rm 2},
        \textbf{Ji Zhang}\textsuperscript{\rm 2}, \textbf{Fei Huang}\textsuperscript{\rm 2}
         \\ \textsuperscript{\rm 1}School of Computer Science and Engineering, Sun Yat-sen University, China\\ \textsuperscript{\rm 2}Alibaba Group \\
         shenwzh3@mail2.sysu.edu.cn, quanxj3@mail.sysu.edu.cn, \\
         ym119608@alibaba-inc.com \\
         \url{https://github.com/X-PLUG/Multi-LLM-Agent}}

\begin{document}
\maketitle
\begin{abstract}
Large Language Model (LLM) agents significantly extend the capabilities of standalone LLMs, empowering them to interact with external tools (e.g., APIs, functions) and complete various tasks in a self-directed fashion. The challenge of tool use demands that LLMs not only understand user queries and generate answers accurately but also excel in task planning, tool invocation, and result summarization. While traditional works focus on training a single LLM with all these capabilities, performance limitations become apparent, particularly with smaller models. To overcome these challenges, we propose a novel approach that decomposes the aforementioned capabilities into a planner, caller, and summarizer. Each component is implemented by a single LLM that focuses on a specific capability and collaborates with others to accomplish the task. This modular framework facilitates individual updates and the potential use of smaller LLMs for building each capability.
To effectively train this framework, we introduce a two-stage training paradigm. First, we fine-tune a backbone LLM on the entire dataset without discriminating sub-tasks, providing the model with a comprehensive understanding of the task. Second, the fine-tuned LLM is used to instantiate the planner, caller, and summarizer respectively, which are continually fine-tuned on respective sub-tasks. Evaluation across various tool-use benchmarks illustrates that our proposed multi-LLM framework surpasses the traditional single-LLM approach, highlighting its efficacy and advantages in tool learning.
\end{abstract}

\begin{figure}[t]
    \centering
    \includegraphics[width=0.9\linewidth]{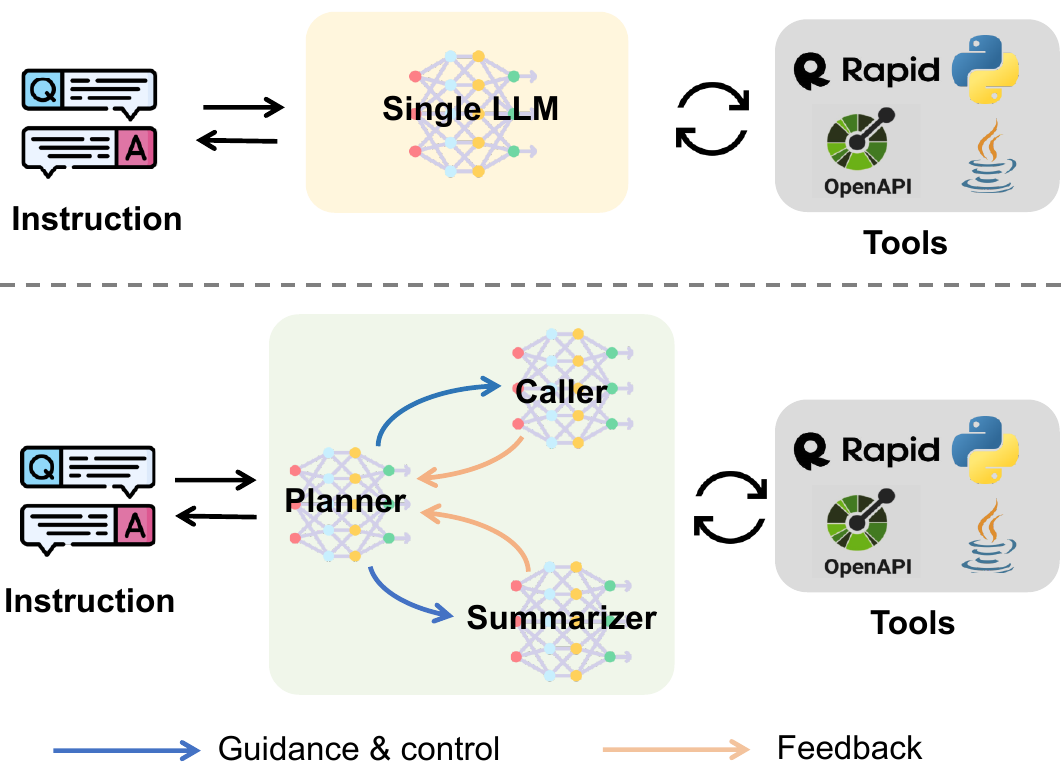}
	\caption{A conceptual comparison of  the traditional single-LLM agent framework (top) and the proposed multi-LLM agent framework, $\alpha$-UMi (bottom). }
	\label{fig:concept}
	\vspace{-0.3cm}
\end{figure}

\section{Introduction}

Large Language Models (LLMs) have revolutionized natural language processing with remarkable proficiency in understanding and generating text. Despite their impressive capabilities, LLMs are not without limitations.~Notably, they lack domain specificity, real-time information, and face challenges in solving specialized problems such as mathematics \citep{gou2023tora} and program compilation~\citep{code_interpertor}. Hence, integrating LLMs with external tools, such as API calls and Python functions, becomes imperative to extend their capabilities and enhance the overall performance. Consequently, LLM agents have become a prominent area for both academia and industry, employing large language models to determine when and how to utilize external tools to tackle various tasks.

In addition to exploring proprietary LLMs like GPT-4, researchers have also actively engaged in developing customizable agent systems by fine-tuning open-source LLMs on diverse tool-use datasets~\citep{gorilla,toolalpaca,qin2023toolllm,gou2023tora}. 
The challenge of tool learning demands sufficiently large and complex LLMs. These models must not only comprehend user queries but also excel in task planning, tool selection and invocation, and result summarization~\citep{tool_learn_survey}. These capabilities draw upon different facets of the LLMs; for instance, planning relies more on reasoning ability, while tool selection and invocation demand legal and accurate request writing, and result summarization requires adept conclusion-drawing skills.
While conventional approaches~\citep{qin2023toolllm, gou2023tora, zeng2023agenttuning}  focus on training a single open-source LLM with all these capabilities, notable performance limitations have been observed, especially with smaller open-source LLMs ~\cite{touvron2023llama, llama2}. Moreover, the tools could be
updated frequently in practical scenarios, when the
entire LLM requires potential retraining.

\begin{figure*}[t]
    \centering
    \includegraphics[width=\textwidth]{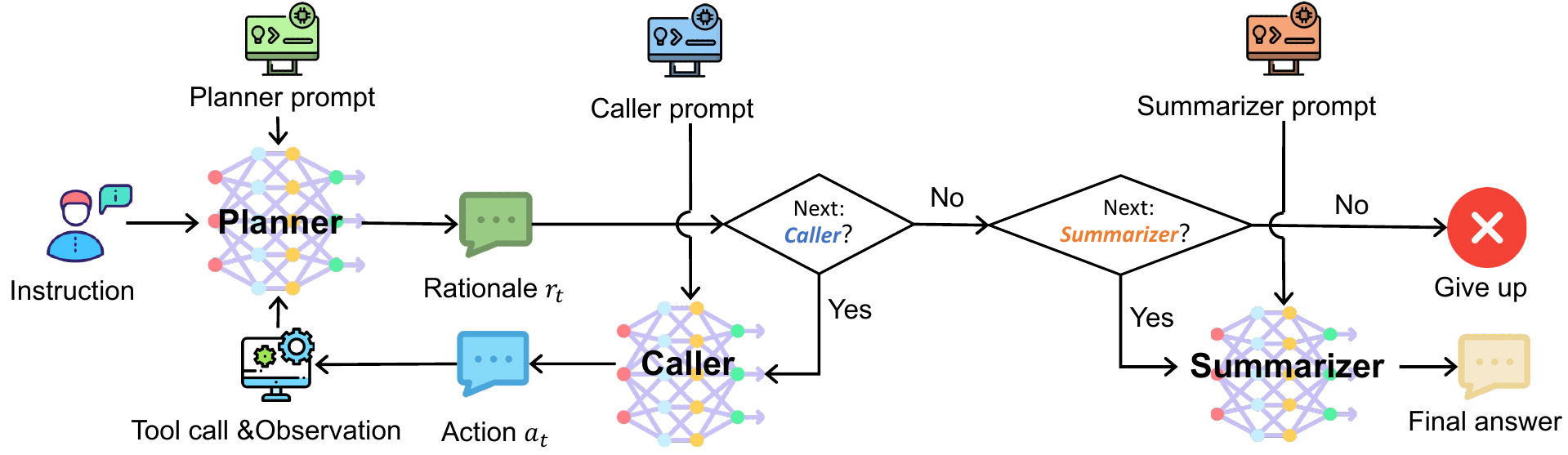}
	\caption{An illustration of how $\alpha$-UMi works to complete a task.}
	\label{fig:framework}
\end{figure*}

To address these challenges, we propose a multi-LLM agent framework for tool learning, $\alpha$-UMi\footnote{In astronomy, the name ``$\alpha$-UMi'' is an alias of the Polaris Star (\url{https://en.wikipedia.org/wiki/Polaris}), which is actually a triple star system consisting of a brighter star (corresponding to the planner) and two fainter stars (corresponding to the caller and the summarizer).}. As illustrated in Figure~\ref{fig:concept}, $\alpha$-UMi decomposes the capabilities of a single LLM into three components, namely planner, caller, and summarizer. Each of these components is implemented by a single LLM and trained to focus on a specific capability. The planner is designed to generate the rationale based on the current state of the system and weighs between selecting the caller or summarizer to generate downstream output, or even deciding to terminate the execution. The caller is directed by the rationale and responsible for invocating specific tools. The summarizer is guided by the planner to craft the ultimate user answer based on the execution trajectory. These components collaborate seamlessly to accomplish various tasks. Compared to previous approaches, our modular framework has three distinct advantages. First, each component undergoes training for a designated role, ensuring enhanced performance for each capability. Second, the modular structure allows for individual updates to each component as required, ensuring adaptability and streamlined maintenance. Third, since each component focuses solely on a specific capability, potentially smaller LLMs can be employed.

To effectively train this multi-LLM framework, we introduce a novel global-to-local progressive fine-tuning strategy (GLPFT). First, an LLM backbone is trained on the original training dataset without discriminating between sub-tasks, enhancing the comprehensive understanding of the tool-learning task. Three copies of this LLM backbone are created to instantiate the planner, caller, and summarizer, respectively. In the second stage, the training dataset is reorganized into new datasets tailored to each LLM's role in tool use, and continual fine-tuning of the planner, caller, and summarizer is performed on their respective datasets.

We employ LLaMA-2~\cite{llama2} series to implement the LLM backbone and evaluate our $\alpha$-UMi agent on several tool learning benchmarks~\cite{qin2023toolllm, toolalpaca}. Experimental results demonstrate that our proposed framework outperforms the single-LLM approach across various model and data sizes. Moreover, we show the necessity of the GLPFT strategy for the success of our framework and delve into the reasons behind the improved performance. Finally, the results confirm our assumption that smaller LLMs can be used in our multi-LLM framework to cultivate individual tool learning capabilities and attain a competitive overall performance.

To sum up, this work makes three critical contributions. First, we demonstrate that small LLMs are weak tool learners and introduce $\alpha$-UMi, a multi-LLM framework for building LLM agents, that outperforms the existing single-LLM approach in tool use. Second, we propose a GLPFT fine-tuning strategy, which has proven to be essential for the success of our framework. Third, we perform a thorough analysis, delving into data scaling laws and investigating the underlying reasons behind the superior performance of our framework.

\section{Related Works}

\subsection{Tool Learning}\label{sec:relate_tool} 
The ability of LLMs to use external tools has become a pivotal component in the development of AI agents, attracting rapidly growing attention~\citep{qin2023toolllm,schick2023toolformer,yang2023gpt4tools,hugginggpt,gorilla,qin2023tool}. Toolformer~\citep{schick2023toolformer} was one of the pioneering work in tool learning, employing tools in a self-supervised manner. Subsequently, a diverse array of external tools has been employed to enhance LLMs in various ways, including the improvement of real-time factual knowledge~\citep{yang2023chatgpt,nakano2021webgpt}, multimodal comprehension and generation~\citep{yang2023gpt4tools,wu2023visual,yang2023mm}, code and math reasoning~\citep{gou2023tora,code_interpertor}, and domain knowledge of specific AI models and APIs~\citep{hugginggpt,li2023modelscopeagent,qin2023toolllm}.
Different from previous approaches relying on a single LLM for tool learning, we introduce a novel multi-LLM collaborated tool learning framework designed for smaller open-source LLMs. This framework decomposes the comprehensive abilities of LLMs into distinct roles, namely a planner, caller, and summarizer.

\subsection{LLM-powered Agents} 
Leveraging the capabilities of LLMs such as ChatGPT~\cite{chatgpt} and GPT-4~\cite{openai2023gpt4}, AI agent systems have found application in diverse scenarios. For instance, solutions like BabyAGI~\cite{babyagi} and AutoGPT~\cite{autogpt} have been developed to address daily problems, while Voyager~\cite{wang2023voyager} and Ghost~\cite{zhu2023ghost} engage in free exploration within Minecraft games. Additionally, MetaGPT \cite{hong2023metagpt}, ChatDev~\cite{qian2023communicative}, and AutoGen~\cite{wu2023autogen} contribute to the development of multi-agent frameworks tailored for software development and problem-solving.

Various techniques have been proposed to augment agent capabilities from different perspectives. The chain-of-thought series~\cite{cot, self-consistant, yao2022react, tot} and Reflextion~\cite{shinn2023reflexion} contribute to enhancing agents' reasoning abilities, while MemoryBank~\cite{zhong2023memorybank} enriches agents with long-term memory. Recent efforts have also emerged in fine-tuning open-source LLMs as agents, exemplified by works like FiREACT~\cite{chen2023fireact} and AgentTuning~\cite{zeng2023agenttuning}. However, these endeavors primarily focus on fine-tuning a single LLM, unlike our approach, which explores an effective method for fine-tuning a multi-LLM agent specifically for tool learning.

\section{Methodology}

\subsection{Preliminary}\label{sec:prime}

Agents for tool learning are systems designed to assist users in completing tasks through a series of decision-making processes and tool use~\cite{tool_learn_survey}. In recent years, these agents commonly adhere to the ReACT framework~\cite{yao2022react}. The backbone of the agent is an LLM denoted as $\mathcal{M}$. Given the user instruction $q$ and the system prompt $\mathcal{P}$, the agent solves the instruction step by step. In the $t$th step, the LLM $\mathcal{M}$ generates a rationale $r_t$ and an action $a_t$ based on the instruction and the current state of the system:
\begin{equation}
    r_t, a_t = \mathcal{M}(\mathcal{P}, \tau_{t-1}, q),
\end{equation}
where $\tau_{t-1}=\{r_1,a_1,o_1,...,r_{t-1},a_{t-1},$ $o_{t-1}\}$ denotes the previous execution trajectory. Here, $o_{t}$ denotes the observation returned by tools when the action $a_t$ is supplied. In the final step of the interaction, the agent generates rationale $r_n$ indicating that the instruction $q$ is solved along with the final answer $a_n$ or that it will abandon this execution run. Therefore, no observation is included in this step.

\subsection{The $\alpha$-UMi Framework}\label{sec:method_framework}

As previously mentioned, the task of tool learning imposes a significant demand on the capabilities of LLMs, including task planning, tool invocation, and result summarization. Coping with all these capabilities using a single open-source LLM, especially when opting for a smaller LLM, appears to be challenging. To address this challenge, we introduce the $\alpha$-UMi framework, which breaks down the tool learning task into three sub-tasks and assigns each sub-task to a dedicated LLM. Figure~\ref{fig:concept} presents an illustration of our framework, which incorporates three distinct LLM components: planner $\mathcal{M}_\text{plan}$, caller $\mathcal{M}_\text{call}$, and summarizer $\mathcal{M}_\text{sum}$. These components are differentiated by their roles in tool use, and each component model has a unique task definition, system prompt\footnote{The prompts for each LLM are provided in Appendix~\ref{app:prompt}.}, and model input.

The workflow of $\alpha$-UMi is shown in Figure~\ref{fig:framework}. Upon receiving the user instruction $q$, the planner generates a rationale comprising hints for the this step. This may trigger the caller to engage with the tools and subsequently receive observations from the tools. This iterative planner-caller-tool loop continues until the planner determines that it has gathered sufficient information to resolve the instruction. At this point, the planner transitions to the summarizer to generate the final answer. Alternatively, if the planner deems the instruction unsolvable, it may abandon the execution. 

\textbf{Planner}: The planner assumes responsibility for planning and decision-making, serving as the ``brain'' of our agent framework. Specifically, the model input for the planner comprises the system prompt $\mathcal{P}_\text{plan}$, the user instruction $q$, and the previous execution trajectory $\tau_{t-1}$. Using this input, the planner generates the rationale $r_t$:
\begin{equation}
    r_t = \mathcal{M}_\text{plan}(\mathcal{P}_\text{\text{plan}},\tau_{t-1}, q).
\end{equation}
Following the rationale, the planner generates the decision for the next step: (1) If the decision is ``Next: Caller'', the caller will be activated and an action will be generated for calling tools. (2) If the decision is ``Next: Summarizer'', the summarizer will be activated to generate the final answer for the user, and the agent execution will finish. (3) If the decision is ``Next: Give up'', it means that the user's instruction cannot be solved in the current situation, and the system will be terminated.



\textbf{Caller:} Interacting with the tools requires the LLM to generate legal and useful requests, which may conflict with other abilities such as reasoning and general response generation during fine-tuning. Therefore, we train a specialized caller to generate the action for using tools. The caller takes the user instruction $q$ and the previous execution trajectory $\tau_{t-1}$ as input. To make the caller focus on the planner's rationale $r_t$ in the current step, we also design a prompt $\mathcal{P}_\text{call}$ to explicitly remind the caller:
\begin{equation}
    a_t = \mathcal{M}_\text{call}(\mathcal{P}_\text{call}, \tau_{t-1}, q, r_t).
\end{equation}

\textbf{Summarizer:} The agent's final response, which aims to offer informative and helpful information to the user, is distinct from the rationales that primarily focus on planning and reasoning. Therefore, we employ a dedicated summarizer tasked with generating the final answer $a_n$. This model utilizes a concise prompt $\mathcal{P}_\text{sum}$ designed to guide the model in concentrating on summarizing the execution trajectory and presenting the answer to the user:
\begin{equation}
    a_n = \mathcal{M}_\text{sum}(\mathcal{P}_\text{sum}, \tau_{n-1}, q, r_n).
\end{equation}

In Figure~\ref{fig:visual_case_1} and  Figure~\ref{fig:visual_case_2}, we show several cases of our $\alpha$-UMi on downstream tasks.

\begin{figure}[t]
    \centering
    \includegraphics[width=0.98\linewidth]{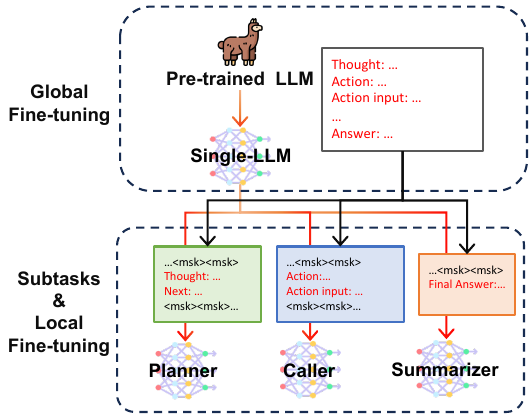}
	\caption{Global-to-local progressive fine-tuning.}
	\label{fig:sft}
	\vspace{-0.3cm}
\end{figure}

\subsection{Global-to-Local Progressive Fine-Tuning}

To effectively fine-tune the above multi-LLM system is a complex endeavor: On one hand, generating the rationale, action, and final answer can facilitate each other during the training process, and enhance the model's comprehension of the entire agent task~\citep{chen2023fireact}. On the other hand, the constraints on model capacity make it challenging to fine-tune a small LLM to achieve peak performance in generating rationales, actions, and final answers simultaneously~\cite{dong2023abilities}. Taking into account these two points, we propose a global-to-local progressive fine-tuning (GLPFT) strategy for $\alpha$-UMi. The motivation behind this strategy is to first exploit the mechanism by which the generation of rationale, action, and final answer can mutually enhance each other. Then, once the single LLM reaches its performance ceiling, it is subsequently split into planner, caller and summarizer for further fine-tuning, in order to enhance its capabilities in the subtasks and mitigate the performance constraints due to limited model capacity.

As depicted in Figure~\ref{fig:sft}, this GLPFT strategy comprises two distinct stages. The first stage involves global fine-tuning, where we fine-tune a backbone LLM on the original training dataset without distinguishing between sub-tasks.~After this stage, the backbone LLM is trained to sequentially output the rationale, action, and answer as introduced in Section~\ref{sec:prime}. 
Then, we create three copies of the backbone LLM, designated as the planner, caller, and summarizer, respectively.

The second stage is local fine-tuning, where we reorganize the training dataset tailored to each LLM's role, as introduced in Section~\ref{sec:method_framework}. We then proceed to fine-tune the planner, caller, and summarizer on their respective datasets, thereby further enhancing their specific abilities in each sub-task. 
During this local fine-tuning stage, we opt to reuse the set of user instructions curated in the global fine-tuning stage. The only adjustment made to the training set is the change in the format of the training data. As illustrated in Figure~\ref{fig:sft}, the fine-tuning objective during the second stage for the planner, caller, and summarizer is oriented towards generating the rationale, action, and final answer, respectively. While the gradients from other text spans are stopped. Simultaneously, we refine the system prompts for the training data of the planner, caller, and summarizer, as detailed in Appendix~\ref{app:prompt}. 

\subsection{Discussions}
The proposed $\alpha$-UMi framework and GLPFT strategy are founded on three main principles. Firstly, the limited ability and capacity of small LLMs, such as LLaMA-7B, pose challenges during fine-tuning in tool learning tasks. In contrast, $\alpha$-UMi decomposes complex tasks into simpler ones, reducing the workload on LLMs. Secondly, $\alpha$-UMi offers increased flexibility in prompt design, allowing us to create specific prompts and model inputs for each LLM to fully leverage its capabilities in sub-tasks. Thirdly, the GLPFT strategy bridges the gap between fine-tuning on the whole tool learning task and on each sub-task, leading to a more successful fine-tuning process of the multi-LLM system. In the following experimental sections, we will focus on demonstrating these principles.


Recent studies have explored multi-agent systems based on LLMs across various domains, such as social communication~\cite{park2023generative,wei2023multi}, software development~\cite{qian2023communicative,hong2023metagpt}, and problem solving~\cite{wu2023autogen}. These frameworks typically rely on powerful closed-source LLMs, demanding advanced capabilities such as automatic cooperation and feedback—abilities that extend beyond those of open-source small LLMs. In contrast, our $\alpha$-UMi aims to ease the LLM's workload in tool-use tasks by integrating multiple LLMs to form an agent, particularly suitable for open-source, small LLMs. Additionally, we introduce the novel GLPFT method for fine-tuning the multi-LLM system, a contribution not found in existing multi-agent works.


\section{Experimental Settings}

\subsection{Benchmarks}


We evaluate the effectiveness of our $\alpha$-UMi on two tool learning benchmarks: ToolBench~\cite{qin2023toolllm} and ToolAlpaca~\cite{toolalpaca}. These tasks involve integrating API calls to accomplish tasks, where the agent must accurately select the appropriate API and compose necessary API requests. Moreover, we partition the test set of ToolBench into in-domain and out-of-domain based on whether the tools used in the test instances have been seen during training. This division allows us to evaluate performance in both in-distribution and out-of-distribution scenarios. For additional details and statistics regarding these datasets, please refer to Appendix~\ref{app:statistic}. We also evaluate $\alpha$-UMi on other benchmarks such as program-aided agent for mathematical reasoning~\cite{math, gsm8k}. The results are shown in Appendix~\ref{app:res_other_benchmark}.

\subsection{Metrics}\label{sec:metrics}
The tasks in ToolBench involve calling APIs through RapidAPI\footnote{https://rapidapi.com/hub.}. This process frequently encounters problems such as API breakdowns, which impacts the fairness of the comparison. To address this problem, we introduce two types of evaluations for ToolBench. In Section~\ref{sec:res_overall}, we first compare the output of agent with the annotated reference at each step\footnote{Refer to Appendix~\ref{app:static_eval} for more details of the evaluation.}, which avoids real-time API callings. The metrics for this evaluation include Action EM (Act. EM), Argument F1 (Arg. F1), and Rouge-L (R-L) as proposed by \citet{li2023modelscopeagent}. 
Moreover, we examine the frequency of API name hallucinations (Hallu.) and the accuracy (Plan ACC) of the agent's planning decisions at each step for using tools invocation, generating answer, or giving up. The reference annotations are based on verified ChatGPT execution results provided in ToolBench. We also provide the results based on real-time RapidAPI calling in Section~\ref{sec:res_toolbench_realtime}, which is the original evaluation method used by the ToolBench team.


For ToolAlpaca, we assess the process correctness rate (Proc.) and the final answer correctness rate (Ans.)~\citep{toolalpaca}, both by GPT-4.

\begin{table*}[t]
	\centering
	\resizebox{\textwidth}{!}{
	\begin{tabular}{l|ccccc|ccccc|cc}
		\toprule
		\multirow{2}*{\textbf{Model}} & \multicolumn{5}{c|}{\textbf{ToolBench (in-domain)}}  & \multicolumn{5}{c|}{\textbf{ToolBench (out-of-domain)}} & \multicolumn{2}{c}{\textbf{ToolAlpaca}}\\ 
		&\textbf{Plan ACC}& \textbf{Act. EM}  &\textbf{Hallu.}& \textbf{Arg. F1} &\textbf{R-L}&\textbf{Plan ACC}& \textbf{Act. EM}  &\textbf{Hallu.}& \textbf{Arg. F1} &\textbf{R-L}&\textbf{Proc.}&\textbf{Ans.}\\
  \hline
        \multicolumn{13}{c}{\textbf{Close-Source LLM}}\\
        \hline 
        ChatGPT (0-shot)& 83.33& 58.67 &7.40 &45.61 &23.08 & 81.62&54.67 & 8.19&40.08& 22.85 & 33&37\\
        GPT-4 (0-shot)& 80.28&55.52&5.98&48.74&28.69 &77.80&55.26&5.12& 47.45& 30.61&41 &44 \\
        \hline
        \multicolumn{13}{c}{\textbf{Model Size = 7B}}\\
        \hline
         ToolLLaMA (len = 4096)& 66.42 & 19.47 & 33.94 & 15.98 & 2.06 & 68.21& 30.75& 25.35& 25.07 &5.78 & -&- \\
         ToolLLaMA (len = 8192)& 77.02 & 47.56 & 4.03 & 42.00 & 15.26 & 77.76& 45.07& 3.45& 40.41 &18.10 & -&- \\
         Single-LLM& 81.92& 53.26& 2.32 &45.57 &\underline{42.66}& 84.61& 56.54& \underline{2.26}& 50.09& \textbf{47.99}& 11& 11 \\

         Multi-LLM$_\text{one-stage}$ & 87.52&45.11	&7.71	&38.02&	41.01	&\underline{88.42}	&53.40	&2.52	&45.79	&\underline{46.39}& 2& 9\\
        Single-LLM$_\text{multi-task}$ & 85.06&51.83&2.96&44.25&27.40&86.55&56.89&2.77&49.50&32.58& \underline{28}& \underline{18} \\
        $\alpha$-UMi$_\text{\emph{w/o} reuse}$ & \underline{88.24}	&\underline{55.50}	&\textbf{0.53}	&\underline{48.97}&	39.98&	87.91&	\underline{58.02}	&2.32	&\underline{50.55}&	42.59 &-&-\\
        $\alpha$-UMi$_\text{\emph{w/} reuse}$ & \textbf{88.92}&	\textbf{58.94}&	\underline{0.57}&	\textbf{52.24}&	\textbf{43.17}&	\textbf{89.72}&	\textbf{60.47}&	\textbf{0.45}&	\textbf{53.60}	&46.26& \textbf{41}& \textbf{35}\\
        \hline
        \multicolumn{13}{c}{\textbf{Model Size = 13B}}\\
        \hline
        Single-LLM & 81.01	&59.67&	1.53&	52.35&	\underline{42.16}	&86.74&	60.04	&\underline{2.03}	&52.94	&\textbf{48.46}&\underline{33}& \underline{29}\\
        Multi-LLM$_\text{one-stage}$ &\underline{86.49}	&50.54	&5.11	&41.96&	36.21&	87.45&	56.71&	3.23&	47.49&	41.62& 22 & 19\\
       Single-LLM$_\text{multi-task}$ &86.36	&58.96&	2.00&	49.28&	28.41&	86.64&	\underline{62.78}	&3.42&	\underline{53.29}	&35.46& 28 & 16\\
        
        $\alpha$-UMi$_\text{\emph{w/o} reuse}$ &86.33	&\underline{60.07}&	\underline{0.39}	&\underline{53.11}&	35.09	&\underline{87.75}	&61.63	&2.95&	52.54	&37.70&- &-\\
         $\alpha$-UMi$_\text{\emph{w/} reuse}$ &\textbf{87.87}&	\textbf{63.03}&	\textbf{0.37}&	\textbf{57.65}&	\textbf{43.46}&	\textbf{88.73}&	\textbf{64.21}	&\textbf{0.24}	&\textbf{57.38}	&\underline{42.50}& \textbf{41}& \textbf{35}\\
		\bottomrule
	\end{tabular}
 }
        \caption{Overall evaluation results on ToolBench and ToolAlpaca. ``ToolLLaMA (len = 4096)'' and ``ToolLLaMA (len = 8192)'' mean setting the max input length of ToolLLaMA to 4096 and 8192, respectively.}
	\label{tab:main_res_toolbench}
	\vspace{-0.3cm}
\end{table*}

\subsection{Implementation Details}\label{sec:implement_detail}
We opt for LLaMA-2-chat-7B/13B~\cite{llama2} as the backbone to implement our framework. In the first stage of our GLPFT, we conduct fine-tuning for the backbone LLM with a learning rate of 5e-5 for 2 epochs. Then, we create three copies of this fine-tuned backbone to instantiate the planner, caller, and summarizer, respectively. In the second stage, we fine-tune the three LLMs with a reduced learning rate of 1e-5. The planner and caller undergo fine-tuning for 1 epoch, while the summarizer undergoes fine-tuning for 2 epochs. We set the global batch size to 48 and employ DeepSpeed ZeRO Stage3~\cite{rajbhandari2021zero} to speed up the fine-tuning process. All experimental results are obtained using greedy decoding, with the maximum sequence length set at 4096.

\subsection{Baselines}
We compare our method with three baseline methods, namely Single-LLM, $\text{Multi-LLM}_{\text{one-stage}}$ and $\text{Single-LLM}_{\text{multi-task}}$. Single-LLM refers to the traditional single-LLM tool learning approach. $\text{Multi-LLM}_{\text{one-stage}}$ involves directly fine-tuning the planner, caller, and summarizer on their own sub-task datasets, without employing our two-stage fine-tuning strategy. $\text{Single-LLM}_{\text{multi-task}}$ refers to using the same LLM to fulfill the roles of planner, caller, and summarizer.  This particular LLM  is fine-tuned on a combined dataset comprising the three sub-task datasets and functions similarly to our multi-LLM framework. We also evaluate the performance of ChatGPT and GPT-4 with 0-shot setting, and ToolLLaMA~\cite{qin2023toolllm}, which is a 7B LLaMA model fine-tuned on ToolBench.

\section{Results and Analysis}
\subsection{Overall Results}\label{sec:res_overall}
The main results are presented in Table~\ref{tab:main_res_toolbench}. We elaborate on our observations from five perspectives:

First, when compared to ChatGPT and ToolLLama, our $\alpha$-UMi outperforms them on all metrics except for the answer correctness on ToolAlpaca. $\alpha$-UMi exceeds these two baselines in terms of Plan ACC and R-L considerably, demonstrating its alignment with annotated reference in terms of planning execution steps and generating final answers. It is worth mentioning that ToolLLaMA only exhibits acceptable performance when the input length is 8192.~At an input length of 4096, ToolLLaMA shows deterioration across various metrics, particularly exhibiting a very high hallucination rate. In contrast, $\alpha$-UMi only requires the input length of 4096 to achieve a satisfying performance. We attribute this to our multi-LLM system design, which allows each small LLM to focus on its sub-task, thereby reducing the requirement for input length. 


Second, $\alpha$-UMi outperforms the Single-LLM agent. On ToolBench, we unveil substantial improvements with $\alpha$-UMi, particularly in Plan ACC, Act. EM, Hallu., and Arg. F1.  On ToolAlpaca, $\alpha$-UMi also surpasses Single-LLM with both 7B and 13B backbones. These findings not only confirm the effectiveness of $\alpha$-UMi in enhancing the agent's planning and API calling capabilities but also suggest a notable decrease in hallucinations, which can significantly elevate user satisfaction.


\begin{table*}[t]
	\centering
	\resizebox{\textwidth}{!}{
	\begin{tabular}{l|l|cc|cc|cc|cc|cc|cc|cc}
		\toprule
		\multirow{2}*{\textbf{Method}}&\multirow{2}*{\textbf{Model}} & \multicolumn{2}{c|}{\textbf{I1-Inst.}}  & \multicolumn{2}{c|}{\textbf{I1-Tool}}& \multicolumn{2}{c|}{\textbf{I1-Cat.}}& \multicolumn{2}{c|}{\textbf{I2-Inst.}}& \multicolumn{2}{c|}{\textbf{I2-Cat.}}& \multicolumn{2}{c|}{\textbf{I3-Inst.}}& \multicolumn{2}{c}{\textbf{Average}}\\ 
		&&Pass&Win&Pass&Win&Pass&Win&Pass&Win&Pass&Win&Pass&Win&Pass&Win\\ 
        \hline
        \multirow{4}*{ReACT} &Claude-2 & 5.5 &31.0 & 3.5 &27.8 &5.5 &33.8 &6.0&35.0 &6.0&31.5 &14.0&47.5 &6.8&34.4  \\
        &ChatGPT& 41.5& - &44.0  &- & 44.5 &- & 42.5 &- & 46.5 &- &22.0  &- &40.2  &- \\
        &ToolLLaMA& 25.0&45.0 &29.0  &42.0 & 33.0 &47.5 & 30.5 &50.8 &31.5  &41.8 & 25.0 &55.0 &29.0  & 47.0\\
        &GPT-4& 53.5&60.0 & 50.0 &58.8 &53.5  &\underline{63.5} &67.0  &   65.8&72.0 &60.3  &47.0 & \underline{78.0} &57.2  &64.4\\
        \hline
        \multirow{7}*{DFSDT} &Claude-2&20.5& 38.0&31.0&44.3 &18.5&43.3 &17.0&36.8 &20.5&33.5 &28.0 &65.0 & 43.1&43.5 \\
        &ChatGPT&54.5 & 60.5& 65.0 &62.0 &60.5 &57.3 &75.0  &72.0 &71.5  &64.8 &62.0  &69.0 &64.8  &64.3 \\
        &ToolLLaMA& 57.0&55.0 & 61.0 &55.3 &62.0 &54.5 &77.0  &68.5 &77.0  &58.0 &66.0  &69.0 &60.7  & 60.0\\
        &GPT-4& 60.0&\textbf{67.5} & \textbf{71.5} & \textbf{67.8}&\textbf{67.0}  &\textbf{66.5} &79.5  & \underline{73.3}&\underline{77.5}  &63.3 &\textbf{71.0}  & \textbf{84.0}&\underline{71.1}  &\textbf{70.4} \\
        &$\alpha$-UMi (7B)& \underline{65.0}&59.5 &68.0  &\underline{66.0} &64.0  & 57.0&\underline{81.5}  &\textbf{76.5} & 76.5 &\underline{72.0} &\underline{70.0}  &63.0 &70.9  &65.9 \\
        &$\alpha$-UMi (13B)& \textbf{65.5}&\underline{61.5} &\underline{69.0}  &\underline{66.0}&\underline{65.0}  & 62.5&\textbf{84.5}  &75.0 & \textbf{81.0} &\textbf{74.5} &\textbf{71.0}  &66.0 &\textbf{72.2}  &\underline{67.7} \\
		\bottomrule
	\end{tabular}
	}
        \caption{Results of real-time evaluation on ToolBench. ``ReACT'' and ``DFSDT'' denote reasoning strategies used to construct agents, as detailed in Section~\ref{sec:res_toolbench_realtime}. ``Win'' measures the relative win rate of each agent compared to ChatGPT-ReACT (``Method''=ReACT, ``Model''=ChatGPT), which does not have an associated win rate.}
	\label{tab:res_toolbench_realtime}
	\vspace{-0.3cm}
\end{table*}

Third, when comparing the results of methods with different model sizes, we note that agents with a 13B backbone exhibit superior performance compared to their 7B counterparts. This observation implies that the shift from a 7B to a 13B model results in a noteworthy improvement in tool utilization capabilities. Significantly, $\alpha$-UMi with a 7B backbone even outperforms the Single-LLM baseline with a 13B LLM, confirming our earlier assumption that smaller LLMs can be utilized in our multi-LLM framework to develop each capability and achieve competitive overall performance.


Fourth, $\alpha$-UMi outperforms $\text{Multi-LLM}_{\text{one-stage}}$ and $\text{Single-LLM}_{\text{multi-task}}$. $\text{Multi-LLM}_{\text{one-stage}}$ even exhibits suboptimal performance compared to the Single-LLM baseline in metrics assessing API calling abilities, such as Act. EM, Hallu., and Arg. F1. This finding highlights the limitations of $\text{Multi-LLM}_{\text{one-stage}}$ when training each LLM on individual sub-tasks, compromising the comprehensive understanding of the tool-use task. Moreover, the subpar performance of $\text{Single-LLM}_{\text{multi-task}}$ reveals a limitation associated with the capacity of 7B and 13B models. The limited model capacity hinders the agent from effectively fulfilling the roles of planner, caller, and summarizer simultaneously. In contrast, through the application of the GLPFT strategy, $\alpha$-UMi successfully mitigates this limitation, showcasing its effectiveness in achieving comprehensive tool learning capabilities.



\begin{figure*}[t]
    \centering
    \includegraphics[width=0.95\textwidth]{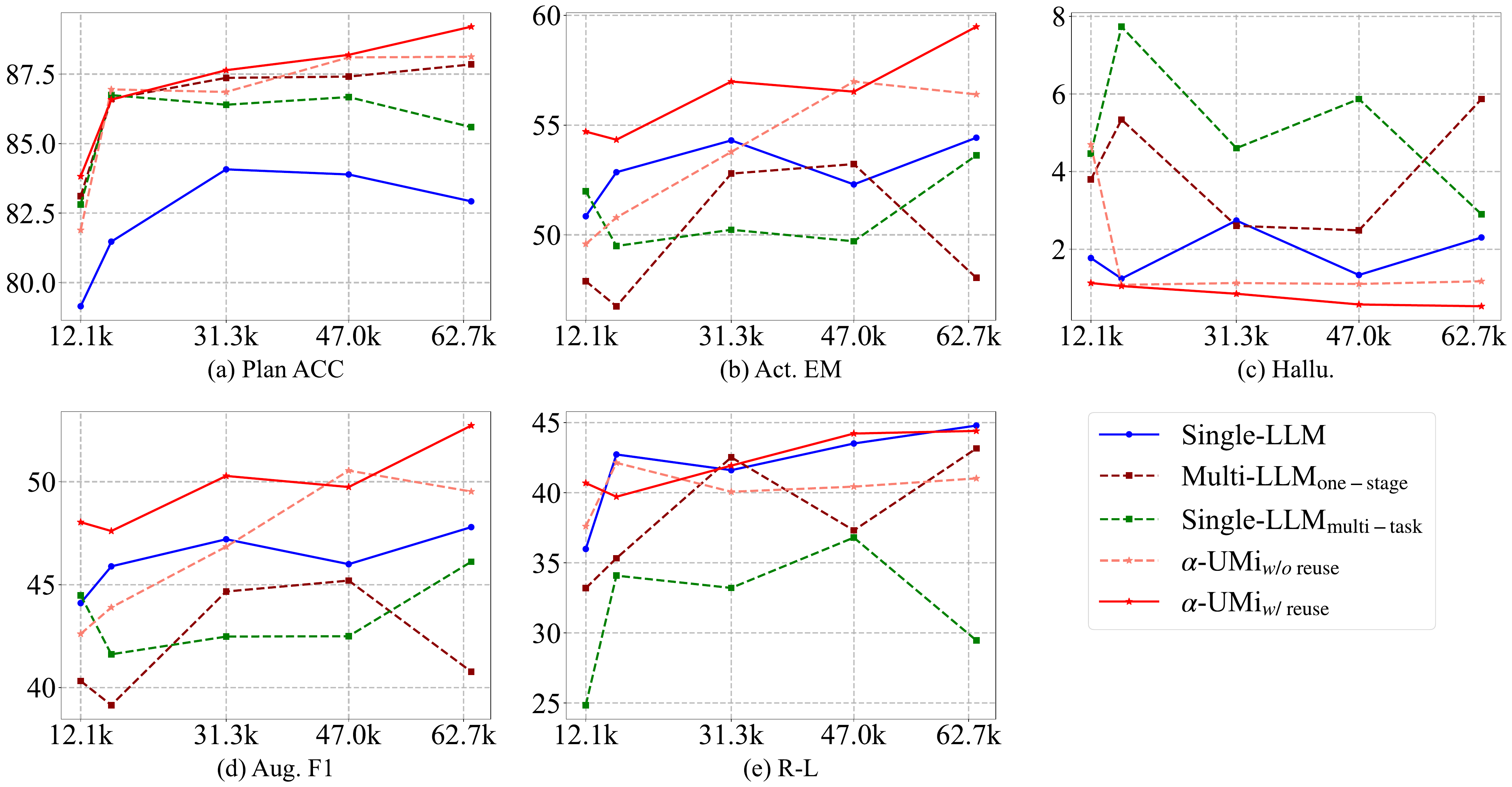}
	\caption{Results of data scaling law study on ToolBench with different evaluation metrics: (a) Plan ACC, (b) Act. EM, (c) Hallu, (d) Arg. F1, and (e) R-L. We randomly sampled five training sets with the scales of 12.1k, 15.7k, 31.3k, 47.0k, and 62.7k instances, accounting for 19.2\%, 25\%, 50\%, 75\%, and 100\% of the training set, respectively.}
	\label{fig:scaling_law}
	\vspace{-0.5cm}
\end{figure*}

Finally, $\alpha$-UMi$_{\text{\emph{w/o} reuse}}$ represents that instead of reusing the user instructions in the first fine-tuning stage of GLPFT, a new set of user instructions are employed for the second stage of GLPFT. This setup is inspired by \citet{flan2}, which has demonstrated that increasing the diversity of user instructions during fine-tuning can improve the performance and generalizability of LLMs. However, as presented in Table~\ref{tab:main_res_toolbench} and visualized in Figure~\ref{fig:scaling_law}, despite the increased diversity of instructions compared to $\alpha$-UMi$_{\text{w/ reuse}}$, $\alpha$-UMi$_{\text{w/o reuse}}$ does not outperform $\alpha$-UMi$_{\text{w/ reuse}}$. We attribute this unexpected result to the following explanation: Since the objectives of the two training stages are different, using distinct sets of user instructions, each with its unique distribution, may introduces a harmful inductive bias that solving one group of the instructions in single-LLM format while the other group in multi-LLM format. In contrast, through the reuse of user instructions, the impact of varying distributions from different sets is mitigated.


\subsection{Real-Time Test on ToolBench}\label{sec:res_toolbench_realtime}

To assess the performance of LLMs for solving real tasks via RapidAPI, we follow the ToolEval method~\cite{qin2023toolllm} proposed by the ToolBench team to conduct a real-time evaluation on the test set of ToolBench. The LLMs under consideration include Claude-2~\cite{claude-2}, ChatGPT, GPT-4, and ToolLLaMA. We apply two reasoning strategies for these LLMs to construct tool learning agents: the ReACT method, as introduced in Section~\ref{sec:prime}, and the Depth First Search-based Decision Tree (DFSDT)~\cite{qin2023toolllm}, which empowers the agent to evaluate and select between different execution paths. Two metrics are included to measure these LLMs' performance: \emph{pass rate}, which calculates the percentage of tasks successfully completed, and \emph{win rate}, which compares the agent's solution path with that of the standard baseline, ChatGPT-ReACT. The above two metrics are assessed by a ChatGPT evaluator with carefully crafted criteria. The empirical results presented in Table~\ref{tab:res_toolbench_realtime} demonstrate that our $\alpha$-UMi (7B) surpasses both ChatGPT and ToolLLaMA by significant margins in terms of \emph{pass rate} (+6.1 and +10.2, respectively) and \emph{win rate} (+1.6 and +5.9, respectively). While $\alpha$-UMi underperforms GPT-4 in \emph{win rate}, it exhibits \emph{pass rates} on par with GPT-4 or even exceeds it in certain test groups such as \emph{I1-Inst.} and \emph{I2-Inst.}. Combining the findings from Section~\ref{sec:res_overall} and this section, we note that our multi-LLM agent outperforms several established baselines across diverse metrics on ToolBench, validating its efficacy.

\subsection{Data Scaling Law}

To assess the impact of the amount of training data on performance, we conduct a data scaling law analysis with the 7B backbone on ToolBench, varying the number of annotated training instances from 12.1k to 62.7k. The results in different metrics are depicted in Figure~\ref{fig:scaling_law}. Several observations can be drawn from the results.
Firstly, when comparing $\alpha$-UMi (solid red curves) to Single-LLM (solid blue curves), there are significant and consistent enhancements in metrics such as Plan ACC, Act. EM, Hallu., and Arg. F1 across various scales of training data. While only minor improvements are observed in the R-L metric, which directly reflects the performance of the summarizer, this suggests that the performance enhancement of our framework is mainly attributed to the separation of the planner and the caller. Secondly, the performances of $\text{Multi-LLM}_{\text{one-stage}}$ and $\text{Single-LLM}_{\text{multi-task}}$ exhibit severe fluctuations in all metrics except for Plan ACC, indicating instability in training the framework through direct fine-tuning or multi-task fine-tuning.
Thirdly, Single-LLM achieves optimal results in different metrics at different data scales. For example, it attains peak performance in Plan ACC with 31.3k instances and the best Arg. F1 and R-L with 62.7k instances. This suggests the challenge of obtaining a single LLM that uniformly performs well across all metrics. In contrast, the performance of our framework consistently improves with increased data scale across all metrics.

\begin{figure}[t]
    \centering
    \includegraphics[width=0.85\linewidth]{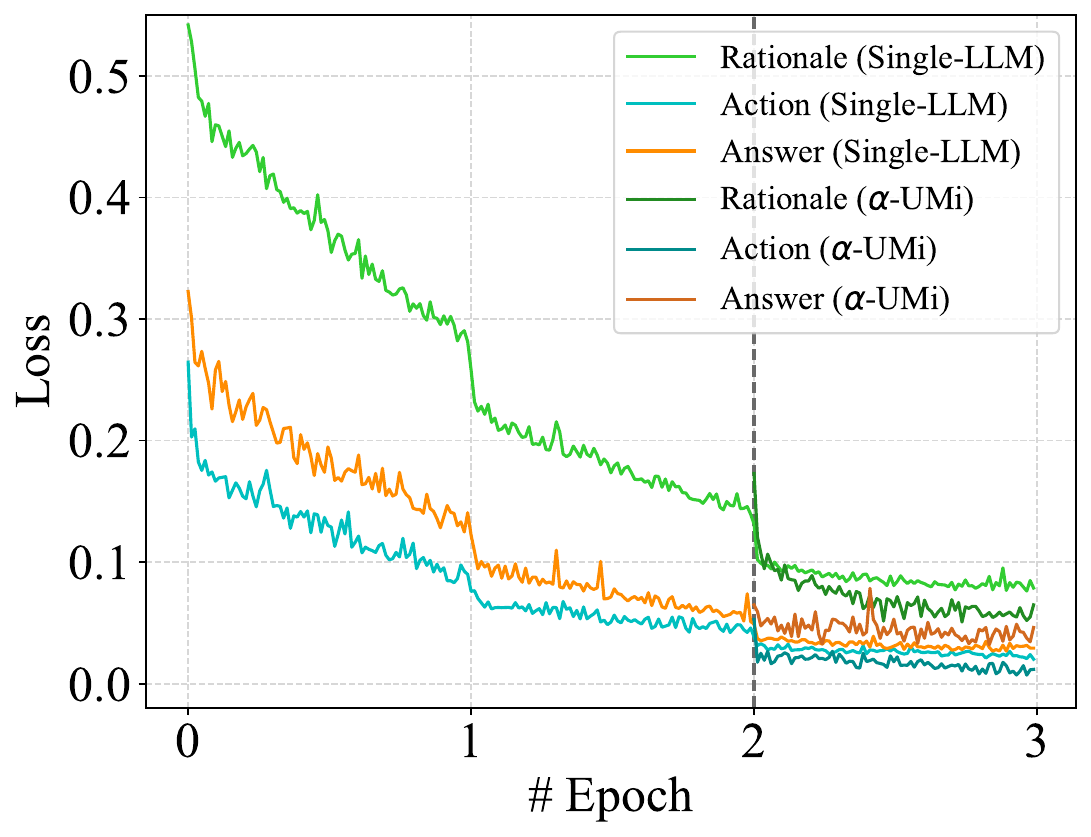}
	\caption{Curves of training loss.}
	\label{fig:training_loss}
	\vspace{-0.5cm}
\end{figure}

\subsection{Why $\alpha$-UMi Works?}

We track the training process of our $\alpha$-UMi approach to examine what makes it different from the Single-LLM baseline. To further investigate how each capability of the model evolves during training, we track the training loss on the rationale, action, and answer components of target responses. The results are depicted in Figure~\ref{fig:training_loss}. As introduced in Section~\ref{sec:implement_detail}, $\alpha$-UMi employs GLPFT and deviates from Single-LLM after two training epochs. Therefore, our discussion focuses on the training curves of $\alpha$-UMi from the third epoch.

The plotted curves reveal a consistent decrease in the training loss for rationale, action, and answer components during the initial two epochs. However, in the third epoch, the losses of Single-LLM exhibit a nearly stagnant trend. In contrast, $\alpha$-UMi experiences continued reductions in the losses associated with rationale and action, indicating further optimization within our $\alpha$-UMi framework.

These observations suggest that the key factor contributing to the success of $\alpha$-UMi lies in its ability to surpass the performance upper-bound of Single-LLM. This is achieved by leveraging GLPFT and decomposing the agent into a multi-LLM system, even after Single-LLM has attained its upper-bound abilities via sufficient fine-tuning.

\subsection{Cost of $\alpha$-UMi}

Given that $\alpha$-UMi operates as a multi-LLM framework, it introduces potential additional costs in terms of training, storage, and deployment. Table~\ref{tab:cost} provides a summary of the costs associated with Single-LLM and $\alpha$-UMi, based on execution logs on 8 Nvidia A100 GPUs with a 40G capacity. Our observations are threefold. Firstly, owing to its composition of a planner, a caller, and a summarizer, $\alpha$-UMi demands three times the storage capacity compared to the Single-LLM framework, assuming they employ backbones of the same size. Secondly, the training of $\alpha$-UMi requires 1.3 times the computational resources and 1.5 times the training duration compared to Single-LLM, while the GPU memory cost for training remains consistent between the two methods. Thirdly, during inference, the time required for both Single-LLM and $\alpha$-UMi is similar, as we only distribute sub-tasks (rationale, action, and answer) to the three LLMs, without forcing them to generate extra contents, thus bringing nearly no extra cost when inference.

\begin{table}[t]
	\centering
	\resizebox{\linewidth}{!}{
	\begin{tabular}{l|c|ccc|c}
		\toprule
		\multirow{2}*{\textbf{Model}} &\multirow{2}*{\textbf{Storage}} &\multicolumn{3}{c|}{\textbf{Train}} & \textbf{Infer.} \\ 
		& & \textbf{Flops} &\textbf{Time} &\textbf{GPU Mem.}& \textbf{Time (Per Inst.)}\\ 
    \hline
    \multicolumn{6}{c}{\textbf{Model Size = 7B}}\\
        \hline
         Single-LLM &7B &$4.8*10^{15}$ &41.54h & 206G& 6.41s \\
         $\alpha$-UMi &7B*3 &$6.2*10^{15}$ &63.34h & 206G & 6.27s  \\
        \hline
    \multicolumn{6}{c}{\textbf{Model Size = 13B}}\\
        \hline
        Single-LLM & 13B&$7.2*10^{15}$ &89.56h & 308G&11.91s  \\
         $\alpha$-UMi & 13B*3& $9.7*10^{15}$&129.96h &  308G& 11.09s \\
		\bottomrule
	\end{tabular}
	}
        \caption{The cost of training and inference.}
	\label{tab:cost}
	\vspace{-0.5cm}
\end{table}

Note that based on the findings presented in Table~\ref{tab:main_res_toolbench}, $\alpha$-UMi with a 7B backbone can outperform Single-LLM with a 13B backbone. Furthermore, the cost associated with $\alpha$-UMi featuring a 7B model is lower than that of Single-LLM featuring a 13B model, both in terms of training and inference. This underscores the cost-effectiveness of $\alpha$-UMi as a means to achieve, and even surpass, the performance of a larger LLM.

\subsection{Case Study}

Figure~\ref{fig:visual_case_1} and Figure~\ref{fig:visual_case_2} show two cases of our $\alpha$-UMi executing real tasks in ToolBench. In the case of Figure~\ref{fig:visual_case_1}, the user specifies the available tools in the instructions, making the tool invocation process simpler. The $\alpha$-UMi framework completes the task within two steps through the collaboration of the planner, caller, and summarizer. In the case of Figure~\ref{fig:visual_case_2}, $\alpha$-UMi initially attempts to use the ``video\_for\_simple\_youtube\_search'' to obtain detailed video information at step 0. However, it realizes that this API has broken and cannot be invoked. Therefore, the planner informs the caller to try an alternative API and obtain accurate information. Ultimately, the user's task is successfully resolved.

To further analyze the specific advantages of our $\alpha$-UMi and Single-LLM frameworks in task execution, we have presented some comparative examples of the two frameworks in Tables~\ref{tab:exe_toolbench_multi_1}, \ref{tab:exe_toolbench_single_1}, \ref{tab:exe_toolbench_multi_2}, and \ref{tab:exe_toolbench_single_2}. Tables ~\ref{tab:exe_toolbench_multi_1} and \ref{tab:exe_toolbench_single_1} illustrate simple tasks that require only a single step tool invocation to be completed, in which case both $\alpha$-UMi and Single-LLM can successfully accomplish the tasks. However, in the complex tasks presented in Tables~\ref{tab:exe_toolbench_multi_2} and \ref{tab:exe_toolbench_single_2}, where the tasks require the models to accomplish some composite objectives, $\alpha$-UMi's planner can quickly understand the user's intentions and plan out steps based on the prompts provided by the caller and summarizer. On the other hand, Single-LLM exhibited some behaviors that did not align with the user's intentions during planning, such as invoking APIs that did not match the intent and entering loops in these misaligned APIs, ultimately failing to provide sufficient information to complete the user's instructions. This result indicates that $\alpha$-UMi's decomposing Single-LLM into a planner, caller, and summarizer reduces the burden on the model during reasoning, allowing the planner model to focus solely on understanding the user's intentions and making effective plans, thereby better accomplishing the tasks.

\section{Conclusion}
The objective of this paper is to address the challenge of designing and fine-tuning a single small LLM to acquire the extensive abilities required for a tool learning agent. To this end, we introduce $\alpha$-UMi, a multi-LLM tool learning agent framework that breaks down the tool learning task into three distinct sub-tasks delegated to three small LLMs: planner, caller, and summarizer. Moreover, we propose a global-to-local progressive fine-tuning strategy and demonstrate its effectiveness in training the multi-LLM framework. We evaluate our approach against single-LLM baselines on four tool learning benchmarks, supplemented by various in-depth analyses, including a data scaling law experiment. Our findings highlight the significance of our proposed method, validating that the system's design for decomposing tool learning tasks and the progressive fine-tuning strategy contribute to enhancing the upper-bound ability of a single LLM. Besides, we acknowledge the potential to utilize small LLMs to surpass the capabilities of an agent framework that relies on a single, larger LLM.


\section{Limitations}
While our framework has been demonstrated to outperform the single-LLM framework in tool learning tasks, there are still some limitations to this work. Firstly, there are additional avenues for exploration, such as integrating small LLMs with a powerful closed-source LLM like GPT-4 to create a ``large + small'' collaborative multi-LLM tool learning agent. Secondly, our framework could be further optimized to enhance its flexibility and applicability to a wider range of agent tasks.

\section{Ethical Statement}

The $\alpha$-UMi framework is trained on the public ToolBench and ToolAlpaca benchmarks, with their original purpose being to enhance the tool invocation capabilities of LLMs and improve their performance in assisting users to complete tasks. This framework has not been trained on any data that poses ethical risks. The backbone model it uses, LLaMA-2-chat, has undergone safety alignment. 



\bibliography{custom.bib}

\newpage
\appendix

\section{System prompts}\label{app:prompt}
\subsection{$\mathcal{P}_{\text{plan}}$ for ToolBench and ToolAlpaca}

\noindent\texttt{You have assess to the following apis:\\
\{doc\}\\
The conversation history is:\\
\{history\}\\
You are the assistant to plan what to do next and whether is caller's or conclusion's turn to answer.\\
Answer with a following format:\\
The thought of the next step, followed by Next: caller or conclusion or give up.}

\subsection{$\mathcal{P}_{\text{call}}$ for ToolBench and ToolAlpaca}

\noindent\texttt{You have assess to the following apis:\\
\{doc\}\\
The conversation history is:\\
\{history\}\\
The thought of this step is:\\
\{thought\}\\
Base on the thought make an api call in the following format:\\
Action: the name of api that should be called in this step, should be exactly in 
\{tool\_names\},\\
Action Input: the api call request.}

\subsection{$\mathcal{P}_{\text{sum}}$ for ToolBench and ToolAlpaca}

\noindent\texttt{Make a conclusion based on the conversation history:\\
\{history\}}

\subsection{$\mathcal{P}_{\text{plan}}$ for MATH and GSM8K}

\noindent\texttt{Solve the math problem step by step  by integrating step-by-step reasoning and Python code,\\
The problem is: \{instruction\}\\
The historical execution logs are:\\
\{history\}
You are the assistant to plan what to do next, and shooce caller to generate code or conclusion to answer the problem.\\
Answer with a following format:\\
The thought of the next step, followed by Next: caller or conclusion.}

\subsection{$\mathcal{P}_{\text{call}}$ for MATH and GSM8K}

\noindent\texttt{The problem is: \{instruction\}\\
The historical execution logs are:\\
\{history\}\\
The thought of this step is:\\
\{thought\}\\
generate the code for this step}

\subsection{$\mathcal{P}_{\text{sum}}$ for MATH and GSM8K}

\noindent\texttt{The problem is: \{instruction\}\\
The historical execution logs are: \\
\{history\}
Make a conclusion based on the conversation history}

\section{Details of Benchmarks}\label{app:statistic}

\subsection{ToolBench}
ToolBench~\cite{qin2023toolllm} is a benchmark for evaluating an agent's ability to call APIs. The ToolBench team collects 16,464 real-world APIs from RapidAPI and a total of 125,387 execution trajectories as the training corpus. We randomly sample 62,694 execution trajectories as the training set, and the average number of execution steps is 4.1. 

The test set of ToolBench is divided into 6 groups, namely I1-instruction, I1-tool, I1-category, I2-instruction, I2-category, and I3-instruction. The groups whose name ends with ``instruction'' means the test instructions in these groups use the tools in the training set, which is the in-domain test data.  Otherwise, the groups whose name ends with ``tool'' or ``category'' means the test instructions do not use the tools in the training set, which is the out-of-domain test data. Each group contains 100 user instructions, therefore the total in-domain test set contains 400 instructions, while the out-of-domain test set contains 200 instructions.

The original evaluation metrics in ToolBench are the pass rate and win rate judged by ChatGPT. However, as introduced in Section~\ref{sec:metrics}, the APIs in RapidAPI update every day, which can cause network block, API breakdown, and exhausted quota. Therefore, to make a relatively fair comparison, we adopt the idea of Modelscope-Agent~\cite{li2023modelscopeagent} to compare the predictions of our model with the annotated GPT-4 outputs on the step level. Specifically, for the $t$th step, we input the model with the previous trajectory of GPT-4, ask our framework to generate the rationale and action of this step, and then compare the generated rationale and action of this step with the output of GPT-4.

\subsection{ToolAlpaca}
ToolAlpaca is another benchmark for evaluating API calling. Unlike ToolBench, the APIs and API calling results in ToolAlpaca are mocked from ChatGPT by imitating how the real APIs work. The total number of training instances in ToolAlpaca is 4098, with an average of 2.66 execution steps per instance. The test set of ToolAlpaca contains 100 user instructions. The evaluation of ToolAlpaca is carried out by a simulator where the agent solves the instruction with the tools mocked by ChatGPT. Finally, GPT-4 judges if the execution process of the agent is consistent with the reference process pre-generated by ChatGPT (Proc. correctness) and whether the final answer generated by the agent can solve the user instruction (Ans. correctness).




\section{Static Evaluation on ToolBench}\label{app:static_eval}

The evaluation method for ToolBench introduced in Section~\ref{sec:metrics} is a static approach that assesses the output of the agent at each step individually. Specifically, for each step $t$, given the ground-truth annotation of the previous execution trajectory $\tau^*_{<t}$, the agent generates the rationale $\hat{r}_t$ and action $\hat{a}_t$ for this step:
\begin{equation}
    \hat{r}_t, \hat{a}_t = \text{Agent}(\tau^*_{<t}).
\end{equation}
Then, metrics are computed by comparing the generated $\hat{r}_t$ and $\hat{a}_t$ with the annotated ground-truth rationale $r^*_t$ and action $a^*_t$ for this step:
\begin{equation}
    \text{Metric} = \text{Evaluate}(\hat{r}_t, \hat{a}_t, r^*_t, a^*_t).
\end{equation}

The advantage of this evaluation method is as follows. At each step, the agent only needs to take the previous ground-truth trajectory as input and outputs the current step's rationale and action. This prevents error propagation due to factors such as network blocks, API breakdowns, and exhausted quotas in any particular step, which could affect the fairness of comparison. This evaluation method is an effective complement to real-time evaluation.

\begin{table}[t]
	\centering
	\resizebox{0.8\linewidth}{!}{
	\begin{tabular}{l|cc}
		\toprule
		\multirow{2}*{\textbf{Model}} & \textbf{MATH} & \textbf{GSM8K}\\ 
		&\multicolumn{2}{c}{\textbf{ACC}}\\  
    \hline
    \multicolumn{2}{c}{\textbf{Model Size = 7B}}\\
        \hline
         Single-LLM  &\underline{17.38}& 37.90 \\
        Multi-LLM$_\text{one-stage}$&15.46 & \underline{38.96}\\
        Single-LLM$_\text{multi-task}$ &14.18 & 27.97\\
        $\alpha$-UMi &\textbf{25.60}& \textbf{49.73}\\
        \hline
    \multicolumn{2}{c}{\textbf{Model Size = 13B}}\\
        \hline
        Single-LLM& 20.26 & \underline{44.88}\\
         Multi-LLM$_\text{one-stage}$& \underline{20.32}&44.57\\
        Single-LLM$_\text{multi-task}$& 15.34 &34.79\\
        $\alpha$-UMi &\textbf{28.54} &\textbf{54.20}\\
		\bottomrule
	\end{tabular}
	}
        \caption{Overall results on  MATH and GSM8K.}
	\label{tab:main_res_math}
	\vspace{-0.5cm}
\end{table}

\section{$\alpha$-UMi on Other Benchmarks}\label{app:res_other_benchmark}

\subsection{MATH and GSM8K}

 The MATH~\cite{math} and GSM8K~\cite{gsm8k} benchmarks are originally designed to test the mathematical reasoning ability of LLMs. Following ToRA~\cite{gou2023tora}, we employ a program-aided agent to solve the mathematical problems presented in these datasets. In our scenario, the planner will generate certain rationales and comments to guide the generation of program, the caller will generate program to conduct mathematical calculation, and finally the summarizer will conclude the final answer. Since ToRA has not released its training data, to facilitate the training of our framework, we utilize \texttt{gpt-3.5-turbo-1106} \cite{chatgpt} and \texttt{gpt-4} \cite{openai2023gpt4} to collect execution trajectories in the training set of MATH and GSM8K and filter out the trajectories that do not lead to the correct final answer. Finally, we collect 5536 trajectories from ChatGPT, 573 trajectories from GPT-4 on MATH, and 6213 from ChatGPT on GSM8K.

The test set sizes of MATH and GSM8K are 5000 and 1319, respectively. During testing, we feed our agent with each of the test instructions and execute the agent with a Python code interpreter. We follow the original evaluation methods of MATH and GSM8K to evaluate the performance of the agent with the accuracy of the final answer. As the evaluation results shown in Table~\ref{tab:main_res_math}, our $\alpha$-UMi can still outperform the baselines on MATH and GSM8K, verifying its effectiveness.

\begin{figure*}[t]
    \centering
    \includegraphics[width=0.95\textwidth]{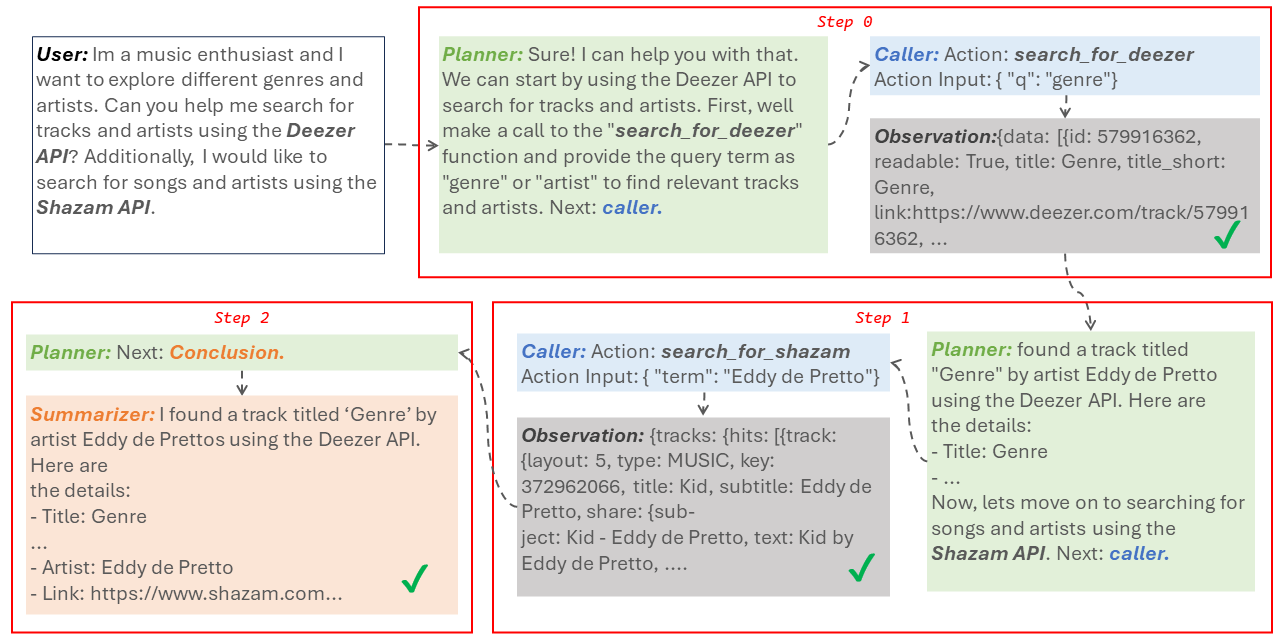}
	\caption{A case study of $\alpha$-UMi. In this case, the user specifies the available tools in the instructions, making the tool invocation process simpler. The $\alpha$-UMi framework completes the task within two steps through the collaboration of the planner, caller, and summarizer.}
	\label{fig:visual_case_1}
	\vspace{-0.5cm}
\end{figure*}

\begin{figure*}[t]
    \centering
    \includegraphics[width=0.95\textwidth]{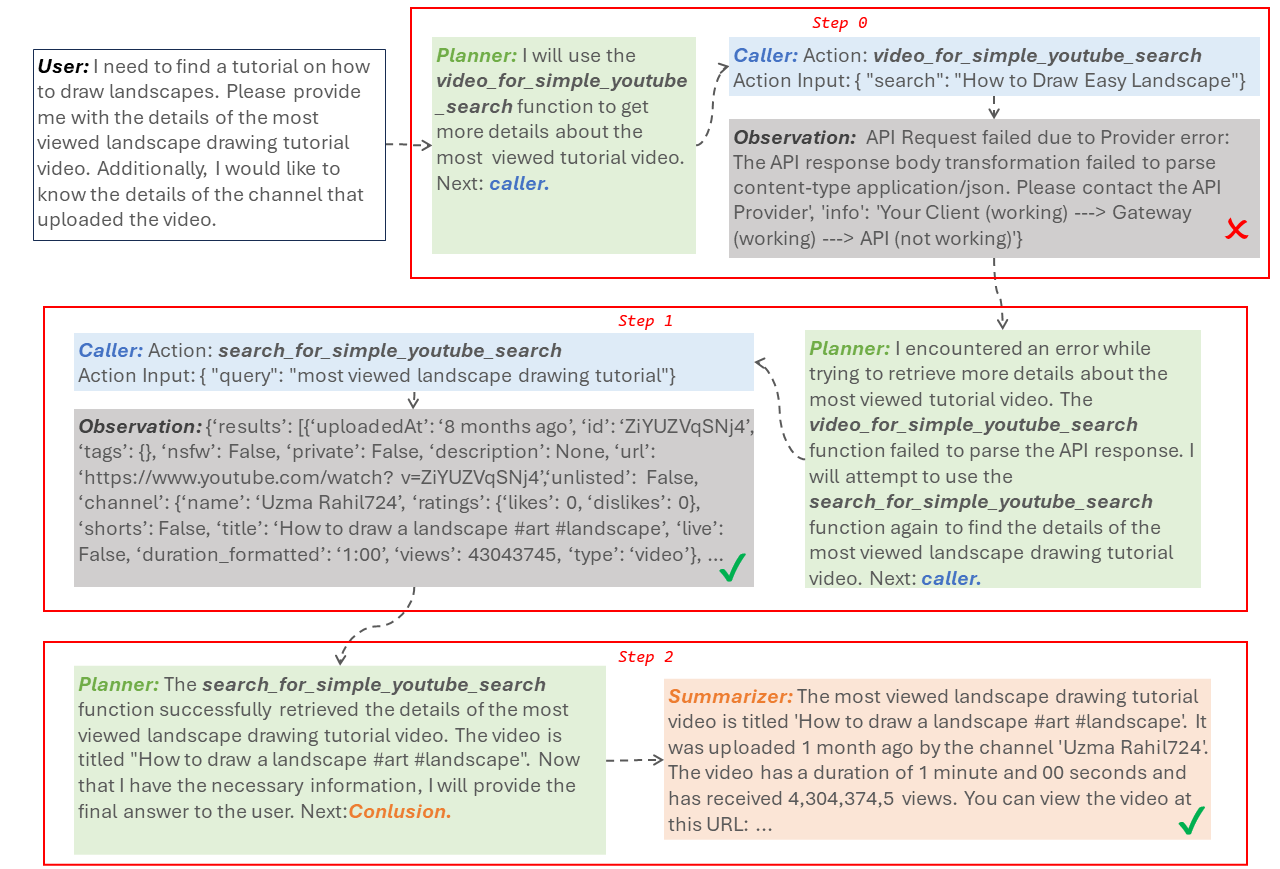}
	\caption{A case study of $\alpha$-UMi, in which $\alpha$-UMi initially attempts to use the ``video\_for\_simple\_youtube\_search'' to obtain detailed video information at step 0. However, it realizes that this API has broken and cannot be invoked. Therefore, the planner informs the caller to try an alternative API and obtain accurate information. Ultimately, the user's task is successfully resolved.}
	\label{fig:visual_case_2}
	\vspace{-0.5cm}
\end{figure*}

\clearpage

\onecolumn
\begin{longtable}{p{65pt}| p{370pt}}
\toprule
\multicolumn{2}{c}{\textbf{Case \#1 of $\alpha$-UMi In ToolBench}} \\ \hline
\multicolumn{2}{p{430pt}}{\textbf{Instruction:} I want to explore the catalog of products available in the 'Electronics' category. Please provide me with the list of products along with their details.}  \\ \hline 
\textbf{Planner}  &Based on the user's request to explore the catalog of products in the 'Electronics' category, I need to call the 'get\_products\_in\_category\_for\_demo\_project\_v13' function with the arguments 'limit': 10, 'skip': 0, and 'category': 'Electronics'. This will allow me to retrieve the list of products along with their details specifically from the 'Electronics' category. By calling this function, I can fulfill the user's request and provide them with the desired information. Next: caller.\\ \hline

\textbf{Caller}  &Action: get\_products\_in\_category\_for\_demo\_project\_v13

Action Input: "\{  "limit": 10, "skip": 0,  "category": "Electronics"\}" \\ \hline

\textbf{Observation}  & \{'products': [\{'id': 893292, 'name': 'TV', 'category': 'electronics'\}, \{'id': 893293, 'name': 'TV', 'category': 'electronics'\}, \{'id': 894481, 'name': 'Gorgeous Plastic Bench', 'category': 'Electronics'\}, \{'id': 896484, 'name': 'Fantastic Rubber Gloves', 'category': 'Electronics'\}, \{'id': 912179, 'name': 'daniel', 'category': 'electronics'\}, \{'id': 916400, 'name': 'Tasty Frozen Pants', 'category': 'Electronics'\}, \{'id': 917106, 'name': 'Tasty Wooden Chips', 'category': 'Electronics'\}, \{'id': 917821, 'name': 'Pizza', 'category': 'Electronics'\}, \{'id': 917875, 'name': 'Tuna', 'category': 'Electronics'\}, \{'id': 917998, 'name': 'Intelligent Frozen Pizza', 'category': 'Electronics'\}], 'summary': \{'count': 105\}\} \\ \hline

\textbf{Planner}  & I have retrieved the list of products available in the 'Electronics' category. Here are some of the products along with their details:

1. Product ID: 893292, Name: TV, Category: Electronics

2. Product ID: 893293, Name: TV, Category: Electronics

3. Product ID: 894481, Name: Gorgeous Plastic Bench, Category: Electronics

4. Product ID: 896484, Name: Fantastic Rubber Gloves, Category: Electronics

5. Product ID: 912179, Name: daniel, Category: Electronics

6. Product ID: 916400, Name: Tasty Frozen Pants, Category: Electronics

7. Product ID: 917106, Name: Tasty Wooden Chips, Category: Electronics

8. Product ID: 917821, Name: Pizza, Category: Electronics

9. Product ID: 917875, Name: Tuna, Category: Electronics

10. Product ID: 917998, Name: Intelligent Frozen Pizza, Category: Electronics

Next: conclusion.
\\ \hline

\textbf{Summarizer}  & 
Here are some of the products available in the 'Electronics' category:

1. Product ID: 893292, Name: TV, Category: Electronics

2. Product ID: 893293, Name: TV, Category: Electronics

3. Product ID: 894481, Name: Gorgeous Plastic Bench, Category: Electronics

4. Product ID: 896484, Name: Fantastic Rubber Gloves, Category: Electronics

5. Product ID: 912179, Name: daniel, Category: Electronics

6. Product ID: 916400, Name: Tasty Frozen Pants, Category: Electronics

7. Product ID: 917106, Name: Tasty Wooden Chips, Category: Electronics

8. Product ID: 917821, Name: Pizza, Category: Electronics

9. Product ID: 917875, Name: Tuna, Category: Electronics

10. Product ID: 917998, Name: Intelligent Frozen Pizza, Category: Electronics
\\ \hline

\hline

\caption{A case of $\alpha$-UMi in ToolBench.}\label{tab:exe_toolbench_multi_1}\\

\end{longtable}

\begin{longtable}{p{65pt}| p{370pt}}
\toprule
\multicolumn{2}{c}{\textbf{Case \#1 of Single-LLM In ToolBench}} \\ \hline
\multicolumn{2}{p{430pt}}{\textbf{Instruction:} I want to explore the catalog of products available in the 'Electronics' category. Please provide me with the list of products along with their details.}  \\ \hline 
\textbf{Assistant}  &Based on the previous actions, it seems that the function "get\_categories\_for\_demo\_project\_v13" was called to retrieve the categories, but it returned an error. To proceed differently, I will call the function "get\_products\_in\_category\_for\_demo\_project\_v13" with the category set to "Electronics" to fetch the products in the Electronics category.

Action: get\_products\_in\_category\_for\_demo\_project\_v13

Action Input: \{
  "limit": 10,
  "skip": 0,
  "category": "Electronics"
\}
\\ \hline

\textbf{Observation}  & \{'products': [\{'id': 893292, 'name': 'TV', 'category': 'electronics'\}, \{'id': 893293, 'name': 'TV', 'category': 'electronics'\}, \{'id': 894481, 'name': 'Gorgeous Plastic Bench', 'category': 'Electronics'\}, \{'id': 896484, 'name': 'Fantastic Rubber Gloves', 'category': 'Electronics'\}, \{'id': 912179, 'name': 'daniel', 'category': 'electronics'\}, \{'id': 916400, 'name': 'Tasty Frozen Pants', 'category': 'Electronics'\}, \{'id': 917106, 'name': 'Tasty Wooden Chips', 'category': 'Electronics'\}, \{'id': 917821, 'name': 'Pizza', 'category': 'Electronics'\}, \{'id': 917875, 'name': 'Tuna', 'category': 'Electronics'\}, \{'id': 917998, 'name': 'Intelligent Frozen Pizza', 'category': 'Electronics'\}], 'summary': \{'count': 105\}\} \\ \hline

\textbf{Assistant}  & I have retrieved the list of products available in the 'Electronics' category. Here are some of the products along with their details:

1. Product ID: 893292, Name: TV, Category: Electronics

2. Product ID: 893293, Name: TV, Category: Electronics

3. Product ID: 894481, Name: Gorgeous Plastic Bench, Category: Electronics

4. Product ID: 896484, Name: Fantastic Rubber Gloves, Category: Electronics

5. Product ID: 912179, Name: daniel, Category: Electronics

6. Product ID: 916400, Name: Tasty Frozen Pants, Category: Electronics

7. Product ID: 917106, Name: Tasty Wooden Chips, Category: Electronics

8. Product ID: 917821, Name: Pizza, Category: Electronics

9. Product ID: 917875, Name: Tuna, Category: Electronics

10. Product ID: 917998, Name: Intelligent Frozen Pizza, Category: Electronics

\\ \hline

\hline

\caption{A case of Single-LLM in ToolBench.}\label{tab:exe_toolbench_single_1}\\

\end{longtable}

\begin{longtable}{p{65pt}| p{370pt}}
\toprule
\multicolumn{2}{c}{\textbf{Case \#2 of $\alpha$-UMi In ToolBench}} \\ \hline
\multicolumn{2}{p{430pt}}{\textbf{Instruction:} I am planning a vacation with my family and we are interested in scuba diving. Can you recommend some dive sites in Thailand? Additionally, we would like to know the top-tier stays available in that area for accommodation options.}  \\ \hline 
\textbf{Planner}  &Based on the user's request, I need to recommend dive sites in Thailand and provide information about the top-tier stays in the area. To accomplish this, I will call the "esites\_by\_a\_country\_or\_a\_region\_for\_world\_scuba\_diving\_sites\_api" function with the argument "country" set to "Thailand". This will allow me to retrieve the relevant dive sites and accommodation options specifically for Thailand. By using this API, I can efficiently gather the necessary information for the user's vacation planning. Next: caller.\\ \hline

\textbf{Caller}  &Action: esites\_by\_a\_country\_or\_a\_region\_for\_world\_scuba\_diving\_sites\_api

Action Input: "\{
  "country": "Thailand"
\}" \\ \hline

\textbf{Observation}  & \{'data': [\{'id': '0126ccc3-6d33-4cb9-a6dd-1519273acf43', 'name': 'The Rock', 'region': 'Racha Noi Island, Phuket', 'lat': '7.552420449639636', 'lng': '98.33433151245119', 'ocean': 'Indian Ocean, Andaman Sea, Strait Of Malacca', 'Location': 'Asia, Thailand, Southern Thailand, Phuket, Racha Noi Island'\}, \{'id': '01c6e49f-45ab-4bb4-bfee-eff9c67f9d22', 'name': 'Paradise Reef', 'region': 'Phuket, Southern Thailand', 'lat': '7.893819147272116', 'lng': '98.2584571838379', 'ocean': 'Indian Ocean, Andaman Sea, Strait Of Malacca', 'Location': 'Asia, Thailand, Southern Thailand, Phuket'\}, \{'id': '02446f8e-9466-4cec-8a7e-940fa2946c44', 'name': 'Shark Fin Reef', 'region': 'Payang Island, Similan Islands', 'lat': '8.50966017107524', 'lng': '97.66388654708864', 'ocean': 'Indian Ocean, Andaman Sea', 'Location': 'Asia, Thailand, Southern Thailand, Phang Nga, Similan Islands, Payang Island'\}, \{'id': '02dea021-ef79-4ee4-929a-46dfb03e5f6e', 'name': 'Bang Bao Reef', 'region': 'Kut Island, Trat', 'lat': '11.612304159956192', 'lng... \\ \hline

\textbf{Planner}  & Based on the user's request, I need to recommend dive sites in Thailand and provide information about the top-tier stays in the area. To accomplish this, I will call the "get\_top\_tier\_stays\_filter\_for\_airbnb\_v2" function. Since the user specifically mentioned the need for top-tier accommodations, I will call this function without any arguments to retrieve the relevant options. By using this API, I can efficiently gather the necessary information for the user's vacation planning and ensure they have a great experience during their trip. Next: caller.
\\ \hline

\textbf{Caller}  &Action: get\_top\_tier\_stays\_filter\_for\_airbnb\_v2

Action Input: "\{\}" \\ \hline

\textbf{Observation}  & \{'status': True, 'message': 'Success', 'timestamp': 1688973421406, 'data': [\{'id': '1', 'title': 'Airbnb Plus', 'subtitle': 'Every Plus home is reviewed for quality'\}, \{'id': '2', 'title': 'Airbnb Luxe', 'subtitle': 'Handpicked luxury homes with personally arranged services'\}]\} \\ \hline

\textbf{Planner}  & The user has requested recommendations for dive sites in Thailand and information about top-tier stays in the area. To fulfill their request, I called the "esites\_by\_a\_country\_or\_a\_region\_for\_world\_scuba\_diving\_sites\_api" function with the argument "country" set to "Thailand". This API allows me to efficiently retrieve relevant dive sites and accommodation options specifically for Thailand. By using this function, I can provide the user with accurate and up-to-date information for their vacation planning. Additionally, I called the "get\_top\_tier\_stays\_filter\_for\_airbnb\_v2" function to retrieve the available options for top-tier accommodations. By using these APIs, I can ensure that the user has a great experience during their trip. Next: conclusion.
\\ \hline

\textbf{Summarizer}  & 
Here are some dive sites in Thailand: The Rock in Racha Noi Island, Phuket; Paradise Reef in Phuket, Southern Thailand; Shark Fin Reef in Payang Island, Similan Islands; Bang Bao Reef in Kut Island, Trat. For accommodation options, you can consider Airbnb Plus and Airbnb Luxe, which offer high-quality and luxury homes respectively. Enjoy your scuba diving adventures in Thailand!
\\ \hline

\hline

\caption{A case of $\alpha$-UMi in ToolBench.}\label{tab:exe_toolbench_multi_2}\\

\end{longtable}

\begin{longtable}{p{65pt}| p{370pt}}
\toprule
\multicolumn{2}{c}{\textbf{Case \#2 of Single-LLM In ToolBench}} \\ \hline
\multicolumn{2}{p{430pt}}{\textbf{Instruction:} I am planning a vacation with my family and we are interested in scuba diving. Can you recommend some dive sites in Thailand? Additionally, we would like to know the top-tier stays available in that area for accommodation options.}  \\ \hline 
\textbf{Assistant}  &Based on the previous action, it seems that the API call to get dive sites in Thailand was not successful. To generate a different action, I will try using the "ranked\_world\_crime\_cities\_for\_ranked\_crime\_cities" function to get a list of ranked crime cities. This will help me identify if Thailand is listed as a crime-free city.

Action:ranked\_world\_crime\_cities\_for\_ranked\_crime\_cities

Action Input: \{\}
\\ \hline

\textbf{Observation}  & \{'products': [\{'id': 893292, 'name': 'TV', 'category': 'electronics'\}, \{'id': 893293, 'name': 'TV', 'category': 'electronics'\}, \{'id': 894481, 'name': 'Gorgeous Plastic Bench', 'category': 'Electronics'\}, \{'id': 896484, 'name': 'Fantastic Rubber Gloves', 'category': 'Electronics'\}, \{'id': 912179, 'name': 'daniel', 'category': 'electronics'\}, \{'id': 916400, 'name': 'Tasty Frozen Pants', 'category': 'Electronics'\}, \{'id': 917106, 'name': 'Tasty Wooden Chips', 'category': 'Electronics'\}, \{'id': 917821, 'name': 'Pizza', 'category': 'Electronics'\}, \{'id': 917875, 'name': 'Tuna', 'category': 'Electronics'\}, \{'id': 917998, 'name': 'Intelligent Frozen Pizza', 'category': 'Electronics'\}], 'summary': \{'count': 105\}\} \\ \hline

\textbf{Assistant}  & Based on the previous action, it seems that the API call to get dive sites in Thailand was not successful. To generate a different action, I will try using the "get\_city\_by\_filter\_for\_ranked\_crime\_cities" function to get the details of a specific city in Thailand. This will help me determine if Thailand is listed as a crime-free city.

Action: get\_city\_by\_filter\_for\_ranked\_crime\_cities

Action Input: \{"id":1\}\\ \hline

\textbf{Observation}  & \{"message": "Function executing from my\_tools.Travel.ranked\_crime\_cities.api import get\_city\_by\_filter error... get\_city\_by\_filter() got an unexpected keyword argument 'id'"\} \\ \hline

\textbf{Assistant}  & Based on the current state and the previous actions, it seems that the API calls to get dive sites in Thailand have been unsuccessful. To generate a different action, I will try using the "esites\_by\_a\_country\_or\_a\_region\_for\_world\_scuba\_diving\_sites\_api" function to get a list of dive sites in Thailand. This might provide the desired information about dive sites in the country.

Action: esites\_by\_a\_country\_or\_a\_region\_for\_world\_scuba\_diving\_sites\_api

Action Input: \{"country": "Thailand"\}\\ \hline

\textbf{Observation}  & \{'data': [\{'id': '0126ccc3-6d33-4cb9-a6dd-1519273acf43', 'name': 'The Rock', 'region': 'Racha Noi Island, Phuket', 'lat': '7.552420449639636', 'lng': '98.33433151245119', 'ocean': 'Indian Ocean, Andaman Sea, Strait Of Malacca', 'Location': 'Asia, Thailand, Southern Thailand, Phuket, Racha Noi Island'\}, \{'id': '01c6e49f-45ab-4bb4-bfee-eff9c67f9d22', 'name': 'Paradise Reef', 'region': 'Phuket, Southern Thailand', 'lat': '7.893819147272116', 'lng': '98.2584571838379', 'ocean': 'Indian Ocean, Andaman Sea, Strait Of Malacca', 'Location': 'Asia, Thailand, Southern Thailand, Phuket'\}, \{'id': '02446f8e-9466-4cec-8a7e-940fa2946c44', 'name': 'Shark Fin Reef', 'region': 'Payang Island, Similan Islands', 'lat': '8.50966017107524', 'lng': '97.66388654708864', 'ocean': 'Indian Ocean, Andaman Sea', 'Location': 'Asia, Thailand, Southern Thailand, Phang Nga, Similan Islands, Payang Island'\}, \{'id': '02dea021-ef79-4ee4-929a-46dfb03e5f6e', 'name': 'Bang Bao Reef', 'region': 'Kut Island, Trat', 'lat': '11.612304159956192', 'lng... \\ \hline

\textbf{Assistant}  & Based on the current state and the previous actions, it seems that the API calls to get dive sites in Thailand have been successful. I have retrieved a list of dive sites in Thailand, including The Rock in Racha Noi Island, Phuket, Paradise Reef in Phuket, Southern Thailand, and Shark Fin Reef in Payang Island, Similan Islands. These dive sites offer potential accommodation options for your family vacation. Now, I will proceed to provide you with the final answer.

conclusion:Here are some dive sites in Thailand that you might find interesting:

1. The Rock - Racha Noi Island, Phuket

2. Paradise Reef - Phuket, Southern Thailand

3. Shark Fin Reef - Payang Island, Similan Islands

4. Bang Bao Reef - Kut Island, Trat

These dive sites offer potential accommodation options for your family vacation. Enjoy your scuba diving adventure in Thailand!\\ \hline

\hline

\caption{A case of Single-LLM in ToolBench.}\label{tab:exe_toolbench_single_2}\\

\end{longtable}

\begin{longtable}{p{65pt}| p{370pt}}
\toprule
\multicolumn{2}{c}{\textbf{Case In ToolAlpaca}} \\ \hline
\multicolumn{2}{p{430pt}}{\textbf{Instruction:} I'm moving to a new apartment and need to update my address with various companies. Can you first verify if my new address is valid and then standardize it? Here's the address: 321 Maple Dr, Apt 12C, Houston, TX, 77002.}  \\ \hline 
\textbf{Planner}  &I need to verify the address and then standardize it. 

Next: caller.\\ \hline

\textbf{Caller}  &Action: verifyUSAddress

Action Input: \{"addressLine1": "321 Maple Dr", "addressLine2":"Apt 12C", "city": "Houston", "state": "TX", "zipCode": "77002"\} \\ \hline

\textbf{Observation}  & \{"status":"valid","standardizedAddress":\{"addressLine1":"123 Main St","addressLine2":"Apt 4B","city":"New York","state":"NY","zipCode":"10001"\}\} \\ \hline

\textbf{Planner}  & The address is valid and has been successfully standardized. 

Next: conclusion.
\\ \hline

\textbf{Summarizer}  & The address you provided is valid and has been successfully standardized. The standardized address is 123 Main St, Apt 4B, New York, NY, 10001
\\ \hline

\hline

\caption{A case study of $\alpha$-UMi in ToolAlpaca.}\label{tab:exe_toolalpaca}\\

\end{longtable}

\begin{longtable}{p{65pt}| p{370pt}}
\toprule
\multicolumn{2}{c}{\textbf{Case In MATH}} \\ \hline
\multicolumn{2}{p{430pt}}{\textbf{Instruction:} Compute the range of the function \$f(x) = sqrt\{x**2\}\$}.  \\ \hline 
\textbf{Planner}  &
Next: caller.\\ \hline

\textbf{Caller}&from sympy import symbols, Interval, oo

def function\_range():   

\ \ \ \ x = symbols(`x')

\ \ \ \ f = sqrt(x**2)

\ \ \ \ range\_f = Interval(0, oo)

\ \ \ \ return range\_f

result = function\_range()

print(result)   \\ \hline

\textbf{Observation}  & NameError: name 'sqrt' is not defined
\\ \hline

\textbf{Planner}  &The code provided for the solution to the given problem will result in an error. To fix this error, the `sqrt` function needs to be imported from the `sympy` library.

Next: caller.\\ \hline

\textbf{Caller} & from sympy import symbols, Interval, oo, sqrt

def function\_range():

\ \ \ \ x = symbols(`x')

\ \ \ \ f = sqrt(x**2)

\ \ \ \ range\_f = Interval(0, oo)

\ \ \ \ return range\_f

result = function\_range()

print(result)\\ \hline

\textbf{Observation}  & Interval(0, oo)
\\ \hline

\textbf{Planner} & Next: conclusion. \\ \hline

\textbf{Summarizer}  & The range of the function \$f(x) = sqrt\{x\**2\}\$ is \$boxed\{[0, infty)\}\$.
\\ \hline

\hline

\caption{A case study of $\alpha$-UMi in MATH.}\label{tab:exe_math}\\

\end{longtable}

\twocolumn

\end{document}